\documentclass[10pt,twocolumn,letterpaper]{article}

\usepackage{iccv}
\usepackage{iccv}
\usepackage{times}
\usepackage{epsfig}
\usepackage{graphicx}
\usepackage{amsmath}
\usepackage{amssymb}

\usepackage{multirow}
\usepackage{subfigure}
\usepackage{algorithmic}
\usepackage{algorithm}

\newtheorem{theorem}{\bf Theorem}
\newtheorem{definition}{\bf Definition}

\usepackage[breaklinks=true,bookmarks=false]{hyperref}

\iccvfinalcopy 

\def\iccvPaperID{****} 

\ificcvfinal\fi

\begin{document}

\title{Deep Multiview Clustering by Contrasting Cluster Assignments}

\author{Jie~Chen\textsuperscript{1},
	Hua~Mao\textsuperscript{2},
	Wai~Lok~Woo\textsuperscript{2},
	~Xi~Peng\textsuperscript{1}\thanks{Corresponding author}\\	
	\textsuperscript{1} College of Computer Science, Sichuan University, China\\
	\textsuperscript{2} Department of Computer and Information Sciences, Northumbria University\\
	{\tt \small chenjie2010@scu.edu.cn; \small\{hua.mao,wailok.woo\}@northumbria.ac.uk;}\\
	{\tt \small pengx.gm@gmail.com}
}

\maketitle
\ificcvfinal\thispagestyle{empty}\fi

\begin{abstract}
   Multiview clustering (MVC) aims to reveal the underlying structure of multiview data by categorizing data samples into clusters. Deep learning-based methods exhibit strong feature learning capabilities on large-scale datasets. For most existing deep MVC methods, exploring the invariant representations of multiple views is still an intractable problem. In this paper, we propose a cross-view contrastive learning (CVCL) method that learns view-invariant representations and produces clustering results by contrasting the cluster assignments among multiple views. Specifically, we first employ deep autoencoders to extract view-dependent features in the pretraining stage. Then, a cluster-level CVCL strategy is presented to explore consistent semantic label information among the multiple views in the fine-tuning stage. Thus, the proposed CVCL method is able to produce more discriminative cluster assignments by virtue of this learning strategy. Moreover, we provide a theoretical analysis of soft cluster assignment alignment. The extensive experimental results obtained on several datasets demonstrate that the proposed CVCL method outperforms several state-of-the-art approaches.
\end{abstract}

\section{Introduction}
Multiview data are usually represented by different types of features or collected from multiple sources. All views share the same semantic information contained in the multiview data. Simultaneously, the data information derived from multiple views is complementary \cite{Chen2022ASR, Xu202SSDFL}. The goal of multiview clustering (MVC) is to divide data samples into different groups according to their distinct feature information.

MVC has attracted increasing attention for many machine learning tasks, including feature selection \cite{Xu202SSDFL}, scene recognition \cite{Tang2022MVSC} and information retrieval \cite{Zhang2022TMVC, Zhang2022MVSC, Chen2022MLRR}. The existing literature involving traditional machine learning techniques can be roughly divided into four categories, including subspace learning-based methods \cite{Chen2022MLRR, Tao2021EC}, nonnegative matrix factorization (NMF)-based methods \cite{Wei2021MCL, Hu2019IMVC}, graph learning-based methods \cite{Chen2022ASR, Li2021IMVC}, and multiple kernel-based methods \cite{Liu2021IMVC, Liu20218LFIMVC}. These traditional shallow models often exhibit limited capabilities to conduct feature representation learning on large-scale datasets \cite{Wang2022MVC}.

A number of deep learning-based methods have been proposed to alleviate the above problems \cite{Wang2022AMCN, Xu2022MLFL, Li2022TCL, Peng2022XAI, Yang2022RMVC, Zhou2020AAN, Li2019DAMVC, Xie2016UDE}. The goal of these deep MVC methods is to learn a more discriminative consensus representation by transforming each view with a corresponding view-specific encoder network. For example, Xie \textit{et al.} \cite{Xie2016UDE} proposed a deep embedding-based clustering method that simultaneously learns feature representations and cluster assignments using deep neural networks. Li \textit{et al.} \cite{Li2019DAMVC} proposed a deep adversarial MVC method that learns the intrinsic structure embedded in multiview data. Zhou \textit{et al.} \cite{Zhou2020AAN} proposed an end-to-end adversarial attention network that makes use of adversarial learning and an attention mechanism to align latent feature distributions and evaluate the importance of different modalities. These methods yield significantly improved clustering performance. However, they fail to consider the semantic label consistency among multiple views, which may lead to difficulty in learning consistent cluster assignments.

Recently, contrastive learning has been integrated into deep learning models to learn discriminative representations of multiple views \cite{Li2021CC, Chen2020CL}. Most existing contrastive learning-based methods attempt to maximize the mutual information contained among the assignment distributions of multiple views \cite{Xu2022MLFL, Wang2020CRL, Caron2020UL}. For example, Yang \textit{et al.} \cite{Yang2022LTNL} took advantage of the available data pairs as positive samples and randomly chose some cross-view samples as negative samples for MVC. In particular, the term “cross-view” means that any two views among multiple views are involved in the contrastive learning process. Caron \textit{et al.} \cite{Caron2020UL} presented an unsupervised visual feature learning method that enforces consistency between the cluster assignments produced for different augmentations. Xu \textit{et al.} \cite{Xu2022MLFL} presented a multilevel feature learning (MFL) method to generate features at different levels for contrastive MVC, e.g., low-level features, high-level features and semantic features. However, which features play critical roles in contrastive feature learning remains unknown \cite{Tian2020GV, Wang2020CRL}. This still leaves an important open question: "which representation should be invariant to multiple views?" Therefore, this motivates us to develop a cross-view contrastive learning (CVCL) model to build a more reasonable view-invariant representation scheme for multiview learning.

In this paper, we present a CVCL method that learns view-invariant representations for MVC. In contrast with most existing deep MVC methods, a cluster-level CVCL strategy is introduced to capture consistent semantic label information across multiple views. By contrasting cluster assignments among multiple views, the proposed CVCL method learns view-invariant representations between positive pairs for MVC. The cluster assignments derived from positive pairs reasonably align the invariant representations among multiple views. Such an alignment flexibly describes the consistent semantic labels obtained from individual views, which are used to measure the intrinsic relationships among the data samples. The $K$-dimensional assignment probability represents the cluster assignment of each sample in the corresponding view. Based on these view-invariant representations, the contrastive loss of the proposed CVCL method encourages the $K$-dimensional cluster assignments produced for positive pairs to be similar and pushes the cluster assignments provided for negative pairs apart. In addition, we provide a theoretical explanation for the realizability of soft cluster assignment alignment.

Our major contributions are summarized as follows.
\begin{itemize}
\item{A CVCL model that contains a two-stage training scheme is proposed to learn view-invariant representations in an end-to-end manner.}
\item{By contrasting the cluster assignments among multiple views, a cluster-level CVCL strategy is presented to explore consistent semantic label information.}
\item{A theoretical analysis of the alignment among the produced view-invariant representations explains why the CVCL model is able to work effectively under certain conditions.}
\item{Extensive experiments conducted on seven multiview datasets demonstrate the effectiveness of the proposed CVCL method.}
\end{itemize}


\section{Related Work}
\label{sec:work}
In this section, we briefly introduce some work related to the proposed CVCL method, including studies on MVC and contrastive learning.

\subsection{Deep Multiview Clustering}
\label{sec:dmvc}
Inspired by recent advances in deep neural network techniques, deep clustering approaches consisting of multiple nonlinear transformations have been extensively studied \cite{Wang2022AMVCN, Xu2022MLFL, Chen2020CL, Zhang2020DPML, Wang2020CRL, Caron2020UL}. One of the representative deep MVC methods is based on deep autoencoders with different regularization terms. These deep autoencoder-based MVC methods aim to learn a consensus representation by minimizing the reconstruction error induced by instances of multiple views. For example, \textit{et al.} \cite{Xie2016UDE} proposed a deep embedded clustering (DEC) method using deep neural networks. DEC first transforms the input sample into features with a nonlinear mapping:
\begin{equation}
{f_\theta }:\mathbf{X} \to \mathbf{Z}
\end{equation}
where $\theta$ is a learnable parameter set. Then the Kullback‒Leibler (KL) divergence between the soft assignment $\mathbf{q}_i$ and the auxiliary distribution $\mathbf{p}_i$ is defined as follows:
\begin{equation}
L = KL\left( {\mathbf{P}||\mathbf{Q}} \right) = \sum\limits_i {\sum\limits_j {{p_{ij}}\log \frac{{{p_{ij}}}}{{{q_{ij}}}}} }.
\end{equation}
The KL divergence loss is minimized to improve the cluster assignment and feature representation effects.

\subsection{Contrastive Learning}
\label{sec:cl}
Contrastive learning has recently achieved significant progress in self-supervised representation learning \cite{Chen2020CL, Tian2020GV, Wang2020CRL}. Contrastive learning-based methods are essentially dependent on a large number of distinct pairwise representation comparisons. Specifically, these methods attempt to maximize the similarities among positive pairs and simultaneously minimize those among negative pairs in a latent feature space. The positive pairs are constructed from the invariant representations of all multiview instances of the same sample. The negative pairs are obtained from the invariant representations of multiple views for different samples. For example, Chen \textit{et al.} \cite{Chen2020CL} presented a visual representation framework for contrastive learning, which maximizes the agreement between differently augmented views of the same example in the latent feature space. Wang \textit{et al.} \cite{Wang2020CRL} investigated the two key properties of the loss function of contrastive learning, i.e., the alignment of features derived from positive pairs and the uniformity of the feature distribution induced on the hypersphere, which can be used to measure the resulting representation quality. These methods are capable of learning good representations based on data argumentation. However, it remains challenging to determine invariant representations for multiple views.

\begin{figure*}[t]
\begin{center}
   \includegraphics[width=0.9\linewidth]{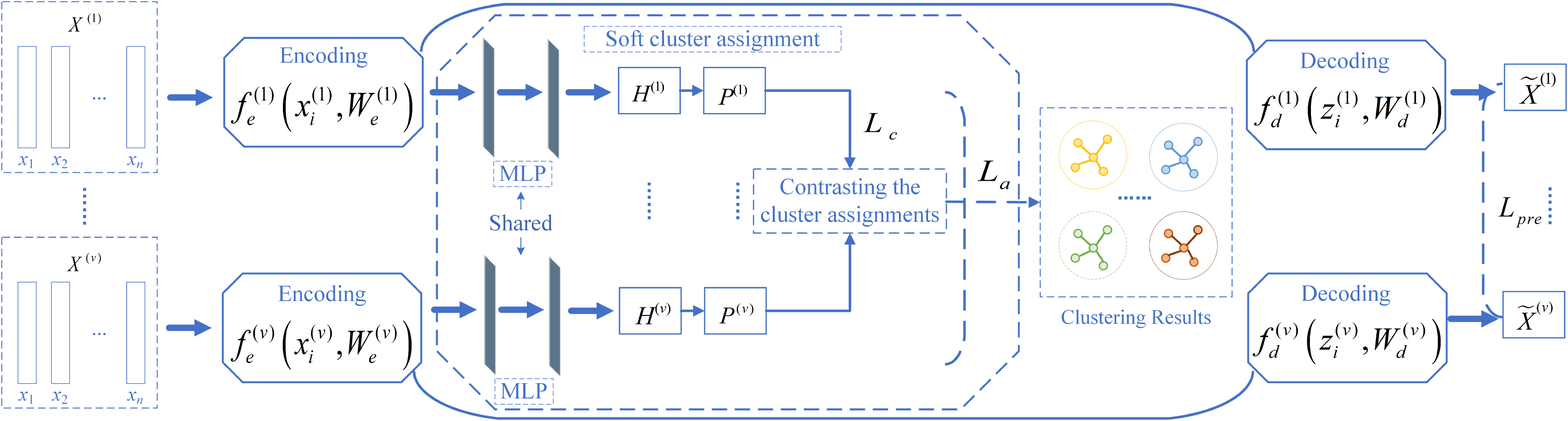}
\end{center}
   \caption{The framework of CVCL. Each view contains two modules, including a view-specific autoencoder module and a CVCL module. The multilayer perceptron (MLP) consists of multiple linear layers. The view-specific autoencoder module contains the encoding part and the decoding part, i.e., $\left\{ {f_e^{(v)}\left( {\mathbf{x}_i^{(v)},\mathbf{W}_e^{(v)}} \right)} \right\}_{v = 1}^{{n_v}}$ and $\left\{ {f_d^{(v)}\left( {\mathbf{z}_i^{(v)},\mathbf{W}_d^{(v)}} \right)} \right\}_{v = 1}^{{n_v}}$, respectively. The CVCL module is employed to explore consistent semantic label information by contrasting the cluster assignments among multiple views.}
\label{fig:architecture}
\end{figure*}

\section{The Proposed Method}
\label{sec:cvcl}
\subsection{Proposed Statement}
\label{sec:se}
Given a set of multiview data $\mathcal{X } = \left\{ {\mathbf{X}^{(v)}} \in {\mathbb{R}^{d_v \times N}} \right\}_{v = 1}^{{n_v}} $ with $n_v$ views and $N$ samples, ${\mathbf{X}^{(v)}}$ represents the $v$th view of the multiview data. Each view ${\mathbf{X}^{(v)}} = \left[ {\mathbf{x}_1^{(v)},\mathbf{x}_2^{(v)},...,\mathbf{x}_N^{(v)}} \right]$  has a total of $N$ instances, where $\mathbf{x}_i^{(v)}$ $\left( {1 \le i \le N} \right)$ represents a $d_v$-dimensional instance. Assume that $K$ is the number of clusters. The samples with the same semantic labels can be grouped into the same cluster. Hence, $N$ samples can be categorized into $K$ different clusters.

\subsection{Network Architecture}
The goal of the proposed CVCL method is to produce semantic labels for end-to-end clustering from the raw instances of multiple views. We introduce an end-to-end deep clustering network architecture by applying contrasting cluster assignments to feature representation learning. As illustrated in Fig. \ref{fig:architecture}, the proposed CVCL network architecture consists of two main modules, i.e., view-specific autoencoder module and cross-view contrastive learning module. The view-specific autoencoder module individually learns clustering-friendly features among multiple views under unsupervised representation learning. The cross-view contrastive learning module achieves the final cluster result by contrasting cluster assignments. With these two modules, CVCL simultaneously learns the view-invariant representations and produce the clustering result for MVC.

\subsection{Cluster-Level CVCL}
Let $f:\mathcal{X }\to \left\{ {{\mathbf{Z}^{(v)}} \in {\mathbb{R}^{N \times k}}} \right\}_{v = 1}^{{n_v}}$ be a function that maps $N$ samples into semantic features. We stack two linear layers and a successive softmax function on the semantic features to produce a cluster assignment probability, which is computed by
\begin{equation}\label{eq:softmax}
\begin{split}
{f_{\left\{ {\mathbf{W}_h^{(v)}} \right\}_{v = 1}^{{n_v}}} }:\left\{ {\mathbf{Z}^{(v)}} \right\}_{v = 1}^{{n_v}} \to \left\{ {\mathbf{H}^{(v)}} \right\}_{v = 1}^{{n_v}}
\end{split}
\end{equation}
where $ \left\{ {\mathbf{W}_h^{(v)}} \right\}_{v = 1}^{{n_v}} $ is a set of learnable parameters.

Inspired by recently proposed contrastive learning techniques, we employ these techniques on the semantic labels to explore the consistency information possessed across multiple views. We can obtain cluster probability matrices $\left\{ {{\mathbf{H}^{(v)}} \in {\mathbb{R}^{N \times K}}} \right\}_{v = 1}^{{n_v}}$ for all views, which are produced on the sematic features of the previous layer. Let $\mathbf{h}^{(v)}_{i}$ be the $i$th row in ${\mathbf{H}^{(v)}}$, and let $h^{(v)}_{ij}$ represent the probability that instance $i$ belongs to cluster $j$ in view $m$. The semantic label of instance $i$ is determined by the largest value among the probabilities in $\mathbf{h}^{(v)}_{i}$.

To increase the differences among the cluster assignments, a unified target distribution $\left\{ {{\mathbf{P}^{(v)}} \in {\mathbb{R}^{N \times K}}} \right\}_{v = 1}^{{n_v}}$  is considered to be a good surrogate for $\left\{ {{\mathbf{H}^{(v)}}} \right\}_{v = 1}^{{n_v}}$, each element of which is calculated as follows:
\begin{equation}\label{eq:td}
\begin{split}
{{p}^{(v)}_{ij}} = \frac{{{{\left(\mathbf{h}_{ij}^{(v)}\right)^2} \mathord{\left/
 {\vphantom {{\left(\mathbf{h}_{ij}^{(v)}\right)^2} {\sum\nolimits_{i = 1}^N {{\mathbf{h}_{ij}}} }}} \right.
 \kern-\nulldelimiterspace} {\sum\nolimits_{i = 1}^N {{\mathbf{h}^{(v)}_{ij}}} }}}}{{\sum\nolimits_{k = 1}^K {\left( {{{\left(\mathbf{h}_{ik}^{(v)}\right)^2} \mathord{\left/
 {\vphantom {{\left(\mathbf{h}_{ik}^{(v)}\right)^2} {\sum\nolimits_{i = 1}^N {{\mathbf{h}_{ik}}} }}} \right.
 \kern-\nulldelimiterspace} {\sum\nolimits_{i = 1}^N {{\mathbf{h}^{(v)}_{ik}}} }}} \right)} }}.
 \end{split}
\end{equation}
Let $\mathbf{p}^{(v)}_{j}$ be the $j$th column of ${\mathbf{P}^{(v)}}$. Each element ${{p}^{(v)}_{ij}}$ in $\mathbf{p}^{(v)}_{j}$ indicates a soft cluster assignment of sample $i$ belonging to cluster $j$. Thus, $\mathbf{p}^{(v)}_{j}$ represents a cluster assignment of the same semantic cluster.

The instances in the different views corresponding to an individual sample share common semantic information. The similarity between two cluster assignments $\mathbf{p}_j^{(v_1)}$ and $\mathbf{p}_j^{(v_2)}$ of cluster $j$ is measured by
\begin{equation}\label{eq:similarity}
\begin{split}
s\left( {\mathbf{p}_j^{(v_1)},\mathbf{p}_j^{(v_2)}} \right) = {\left( {\mathbf{p}_j^{(v_1)}} \right)^T}\mathbf{p}_j^{(v_2)}
\end{split}
\end{equation}
where $v_1$ and $v_2$ denote two distinct views. The cluster assignment probabilities of the instances among different views should be similar in the CVCL module since these instances characterize the same sample. Moreover, the instances in multiple views are irrelevant to each other if they are used to characterize different samples. Therefore, there are $\left( {{n_v} - 1} \right)$ positive cluster assignment pairs and ${n_v}\left( {K - 1} \right)$ negative cluster assignment pairs when considering $\mathbf{p}^{(v)}_{j}$ and $K$ clusters across $n_v$ views.

The similarities among the intracluster assignments should be maximized, and those among the intercluster assignments should be minimized. We simultaneously cluster the samples while enforcing consistency among the cluster assignments. The cross-view contrastive loss between $\mathbf{p}_k^{({v_1})}$ and $\mathbf{p}_k^{({v_2})}$ is defined as follows:

\begin{equation}\label{eq:lossl}
\vspace{-0.4cm}
\begin{split}
& {l^{({v_1},{v_2})}} =-\frac{1}{K}\sum\limits_{k = 1}^K {\log \frac{{{e^{{{s\left( {\mathbf{p}_k^{({v_1})},\mathbf{p}_k^{({v_2})}} \right)} \mathord{\left/
 {\vphantom {{s\left( {\mathbf{p}_k^{({v_1})},\mathbf{p}_k^{({v_2})}} \right)} \tau }} \right.
 \kern-\nulldelimiterspace} \tau }}}}}T}, \\
& T = {{\sum\limits_{j = 1,j \ne k}^K {{e^{{{s\left( {\mathbf{p}_j^{({v_1})},\mathbf{p}_k^{({v_1})}} \right)} \mathord{\left/
 {\vphantom {{s\left( {\mathbf{p}_j^{({v_1})},\mathbf{p}_k^{({v_1})}} \right)} \tau }} \right.
 \kern-\nulldelimiterspace} \tau }}}}  + \sum\limits_{j = 1}^K {{e^{{{s\left( {\mathbf{p}_j^{({v_1})},\mathbf{p}_k^{({v_2})}} \right)} \mathord{\left/
 {\vphantom {{s\left( {\mathbf{p}_j^{({v_1})},\mathbf{p}_k^{({v_2})}} \right)} \tau }} \right.
 \kern-\nulldelimiterspace} \tau }}}} }}
\end{split}
\end{equation}
where $\tau$ is a temperature parameter, $\left( {\mathbf{p}_k^{({v_1})},\mathbf{p}_k^{({v_2})}} \right)$ is a positive cluster assignment pair between two views $v_1$ and $v_2$, and $\left( {\mathbf{p}_j^{({v_1})},\mathbf{p}_k^{({v_1})}} \right)$ $\left( j \ne k \right)$ and $\left( {\mathbf{p}_j^{({v_1})},\mathbf{p}_k^{({v_2})}} \right)$ are the negative cluster assignment pairs in two views $v_1$ and $v_2$, respectively. The cross-view contrastive loss induced across multiple views is designed as:
\begin{equation}\label{eq:lossc}
\begin{split}
{L_c} = \frac{1}{2}\sum\limits_{v_1 = 1}^{{n_v}} {\sum\limits_{v_2 = 1,v_2 \ne v_1}^{{n_v}} {{l^{(v_1,v_2)}}} }.
\end{split}
\end{equation}
The cross-view contrastive loss explicitly compares pairs of cluster assignments among multiple views. It pulls pairs of cluster assignments from the same cluster together and pushes cluster assignments from different clusters away from each other.

To prevent all instances from being assigned to a particular cluster, we introduce a regularization term as follows:
\begin{equation}\label{eq:lossa}
\vspace{-0.2cm}
\begin{split}
{L_a} = \sum\limits_{v = 1}^{{n_v}} {\sum\limits_{j = 1}^K {q_j^{(v)}\log q_j^{(v)}} }
\end{split}
\end{equation}
where $q_j^{(v)}$ is defined as $q_j^{(v)} = \frac{{\sum\limits_{i = 1}^N {p_{ij}^{(v)}} }}{N}$. This term is regarded as a cross-view consistency loss in the CVCL model \cite{Xu2022MLFL}. Assume that all instances belong to a single cluster $j$. This implies that $p_{ij}^{(v)} = 1$ for all $i=1, 2, ..., N$ such that $q_j^{(v)}\log q_j^{(v)} = 0$. As $0 \le p_{ij}^{(v)} \le 1 $, we have $q_j^{(v)}\log q_j^{(v)} \le 0$. This means that the following inequality,
\begin{equation}\label{eq:inequality}
\begin{split}
q_j^{(v)}\ln q_j^{(v)} < 0,
\end{split}
\end{equation}
holds if each cluster has at least one instance. This encourages more elements to reside in each greater-than-zero row of ${\mathbf{P}^{(v)}}$ by minimizing ${L_{a}}$ in Eq. \eqref{eq:lossa}. Therefore, the network is able to encourage cross-view consistency across the cluster assignment probabilities of different instances among the multiple views of each sample with Eq. \eqref{eq:lossa}.

\subsection{Two-Stage Training Scheme}
As illustrated in Fig. \ref{fig:architecture}, we first perform a pretraining task with a deep autoencoder for parameter initialization. Then, we employ a fine-tuning step to train the whole network for MVC.

\subsubsection{Parameter Initialization via a Pretraining Network}
We design a pretraining network that is made up of a view-specific encoder module $f_e^{(v)}$ $\left( {1 \le v \le {n_v}} \right)$ and a corresponding decoder module $f_d^{(v)}$ for each view. The encoder module learns the embedded feature representations by
\begin{equation}\label{eq:encoder}
\begin{split}
\mathbf{z}_i^{(v)} = f_e^{(v)}\left( {\mathbf{x}_i^{(v)},\mathbf{W}_e^{(v)}} \right)
\end{split}
\end{equation}
where $\mathbf{z}_i^{(v)}$ is the embedded feature representation of ${\mathbf{x}_i^{(v)}}$. The decoder module reconstructs the sample ${\mathbf{x}_i^{(v)}}$ as follows
\begin{equation}\label{eq:decoder}
\begin{split}
\mathbf{\widetilde x}_i^{(v)} & = f_d^{(v)}\left( {f_e^{(v)}\left( {\mathbf{x}_i^{(v)},\mathbf{W}_e^{(v)}} \right),\mathbf{W}_d^{(v)}} \right) \\
\end{split}
\end{equation}
where $\mathbf{\widetilde x}_i^{(v)}$ is a reconstruction of $\mathbf{x}_i^{(v)}$. Each encoder or decoder module consists of four or more layers in the proposed CVCL model. The nonlinear rectified linear unit (ReLU) function is chosen as the activation function in the deep autoencoder.

For multiple views, the reconstruction loss of the pretraining network between the input and output is designed as:
\begin{equation}\label{eq:lossr}
\vspace{-0.3cm}
\begin{split}
{L_{pre}} & = \sum\limits_{v = 1}^{{n_v}} {\sum\limits_{i = 1}^N {\left\| {\mathbf{x}_i^{(v)} - f_d^{(v)}\left( {f_e^{(v)}\left( {\mathbf{x}_i^{(v)},\mathbf{W}_e^{(v)}} \right),\mathbf{W}_d^{(v)}} \right)} \right\|_2^2} }
\end{split}
\end{equation}
This is considered to be a pretraining stage for parameter initialization.

\begin{algorithm}
\renewcommand{\algorithmicrequire}{\textbf{Input:}}
\renewcommand\algorithmicensure {\textbf{Output:} }
\caption{Optimization procedure for CVCL}
\label{alg:se}
\begin{algorithmic}
\REQUIRE Data matrices $\left\{ {\mathbf{X}^{(v)}} \right\}_{v = 1}^{{n_v}}$, the numbers of samples $N$ and epochs $epochs$, parameters $\alpha$ and $\beta$.
\end{algorithmic}
\begin{algorithmic}[1]
\STATE Initialize $ \left\{ {\mathbf{W}^{(v)}} \right\}_{v = 1}^{{n_v}} $ by minimizing ${L_{pre}}$ in Eq. \eqref{eq:lossr}; \\
\FOR{ ${t = 1}$ to $epochs$ }
\STATE Choose a random minibatch of samples;
\STATE Computing $\left\{ {{\mathbf{\widetilde X}^{(v)}}} \right\}_{v = 1}^{{n_v}}$ and $\left\{ {{\mathbf{H}^{(v)}}} \right\}_{v = 1}^{{n_v}}$ via Eqs. \eqref{eq:encoder} and \eqref{eq:softmax}, respectively;
\STATE Computing $\left\{ \mathbf{P}^{(v)} \right\}_{v = 1}^{{n_v}}$ via Eq. \eqref{eq:td}; \\
\STATE Computing $ \left\{ {\mathbf{W}^{(v)}} \right\}_{v = 1}^{{n_v}} $ and $ \left\{ {\mathbf{W}_h^{(v)}} \right\}_{v = 1}^{{n_v}} $ by minimizing ${L_{fine}}$ in Eq. \eqref{eq:lossfine}; \\
\ENDFOR
\STATE Calculate semantic labels by Eq. \eqref{eq:label};
\ENSURE The label predictions $\mathbf{Y} = \left[ {{y_1},{y_2},...,{y_n}} \right]$.
\end{algorithmic}
\end{algorithm}

\subsubsection{MVC via a Fine-tuning Network}
The overall loss of the proposed method consists of three main components: the reconstruction loss of the pretraining network, the cross-view contrastive loss and the cross-view consistency loss, i.e.,
\begin{equation}\label{eq:lossfine}
\vspace{-0.3cm}
\begin{split}
{L_{fine}} = {L_{pre}} + \alpha {{L_c} + \beta {L_a}}
\end{split}
\end{equation}
where $\alpha$ and $\beta$ are tradeoff hyperparameters.

The proposed method aims to learn the common semantic labels from the feature representations, which are generated from the instances of multiple views. Let $\mathbf{q}^{(v)}_{i}$ be the $i$th row of ${\mathbf{P}^{(v)}}$, and let ${q_{ij}^{(v)}}$ denote the $j$th element of $\mathbf{q}^{(v)}_{i}$. Specifically, $\mathbf{q}^{(v)}_{i}$  is the $K$-dimensional soft assignment probability, where $\sum\limits_{i = 1}^K {q_{ij}^{(v)} = 1}$. Once the training process of the network is completed, the semantic label of sample $i$ $\left( {1 \le i \le N} \right)$ can be predicted by
\begin{equation}\label{eq:label}
\begin{split}
{y_i} = \mathop {\arg \max }\limits_j \left( {\frac{1}{{{n_v}}}\sum\limits_{v = 1}^{{n_v}} {q_{ij}^{(v)}} } \right).
\end{split}
\end{equation}

An adaptive momentum-based minibatch gradient descent method \cite{Qian1999MT} is employed to optimize the whole network during the two training stages. The final clustering results are produced by the deep autoencoder with the CVCL module. The entire optimization procedure of the proposed method is summarized in Algorithm \ref{alg:se}.

\subsection{Theoretical Analysis}

\subsubsection{Generalization Bound of the Loss Function}
We analyze the generalization bound of the loss function in the proposed method. According to Theorem \ref{th:bound}, ${L_c}$ has a specific lower bound in Eq. \eqref{eq:lossc}. The proof of Theorem \ref{th:bound} can be found in the supplementary material. Assume that each cluster must have at least one multiview data sample. This indicates that $q_{ij}^{(v)} > 0$ in $q_j^{(v)}$. A constant $c$ must exist such that ${q_{ij}^{(v)}\log q_{ij}^{(v)}} > c $ in Eq. \eqref{eq:lossa} holds for all $j=1,2,...,K$. This shows that ${L_a}$ also has a lower bound in Eq. \eqref{eq:lossa}. Therefore, a lower bound is theoretically guaranteed to be obtained when minimizing ${L_{fine}}$ in Eq. \eqref{eq:lossfine}.

\begin{theorem} \label{th:bound}
Assume that there are $N$ samples and $K$ clusters. Given two views $v_1$ and $v_2$ and ${l^{({v_1},{v_2})}}$ in Eq. \eqref{eq:lossl}, the following inequality holds:
\begin{equation*}
{l^{({v_1},{v_2})}} \ge {e^{\log \left( {2K - 1} \right) - {N \mathord{\left/
 {\vphantom {N \tau }} \right.
 \kern-\nulldelimiterspace} \tau }}}.
\end{equation*}
\end{theorem}

\subsubsection{Realizability of Soft Cluster Assignment Alignment}
For the sake of discussion, we assume that there are three clusters with sizes of $k_1$, $k_2$ and $k_3$ and $N$ samples $\left( {N=k_1+k_2+k_3} \right)$. We consider an ideal case in which all instances in different clusters strictly belong to the respective low-dimensional subspaces in each view. Without loss of generality, each ${\mathbf{P}^{(v)}}$ can be represented by
\begin{equation}
{\mathbf{P}^{({v})}} = \left[ {\begin{array}{*{20}{c}}
{\mathbf{P}_1^{({v})}}\\
{\mathbf{P}_2^{({v})}}\\
{\mathbf{P}_3^{({v})}}
\end{array}} \right] = \left[ {\begin{array}{*{20}{c}}
{\overrightarrow {{\mathbf{1}_{{k_1}}}} }&{\overrightarrow {{\mathbf{0}_{{k_1}}}} }&{\overrightarrow {{\mathbf{0}_{{k_1}}}} }\\
{\overrightarrow {{\mathbf{0}_{{k_2}}}} }&{\overrightarrow {{\mathbf{1}_{{k_2}}}} }&{\overrightarrow {{\mathbf{0}_{{k_2}}}} }\\
{\overrightarrow {{0_{{\mathbf{k}_3}}}} }&{\overrightarrow {{\mathbf{0}_{{k_3}}}} }&{\overrightarrow {{\mathbf{1}_{{k_3}}}} }
\end{array}} \right]
\end{equation}
where ${\overrightarrow {{\mathbf{1}_{{k_1}}}} }$ denotes a column vector of all ones with a size of $k_1$. In particular, we always find a matrix transpose $\mathbf{T}$ to obtain such a matrix ${\mathbf{P}^{(v)}}$ if the arrangement assumption is violated. For any two views $v_1$ and $v_2$, ${\mathbf{P}^{(v_1)}}$ is identical to ${\mathbf{P}^{(v_2)}}$. Hence, the $\left\{ {{\mathbf{P}^{(v)}}} \right\}_{v = 1}^{{n_v}}$ are invariant to all types of instances for multiple views.

\begin{definition}[Strict Alignment] \label{def:alignment}
Given an encoder $f$, we have ${\mathbf{P}^{({v_1})}} = f\left( {{\mathbf{X}^{({v_1})}}} \right) \in {\mathbb{R}^{N \times K}}$ and ${\mathbf{P}^{({v_2})}} = f\left( {{\mathbf{X}^{({v_2})}}} \right) \in {\mathbb{R}^{N \times K}}$, where ${\mathbf{X}^{(v_1)}}$ and ${\mathbf{X}^{(v_2)}}$ represent the instances of views $v_1$ and $v_2$, respectively. The encoder $f$ is strictly aligned if the following conditions are satisfied: $\forall {v_1},{v_2} \in \{ 1,2,...{n_v}\} $, ${v_1} \ne {v_2}$; $\forall {i},{j} \in \{ 1,2,...{K}\} $, ${i} \ne {j}$; and $\forall {k} \in \{ 1,2,...{N}\} $,
\begin{equation*}
\begin{split}
& (1) \ \mathbf{p}_i^{({v_1})} = \mathbf{p}_i^{({v_2})};\\
& (2) \ p_{ik}^{({v_1})} = \left\{ {\begin{array}{*{20}{c}}
{1, \text{the } k\text{th sample belongs to the } i\text{th cluster}}\\
{0, \qquad \qquad \qquad \qquad \qquad \qquad otherwise}
\end{array}} \right.; \\
& (3) \ \left\langle {\mathbf{p}_i^{({v_1})},\mathbf{p}_j^{({v_1})}} \right\rangle  = \left\langle {\mathbf{p}_i^{({v_1})},\mathbf{p}_j^{({v_2})}} \right\rangle  = \mathbf{0} \\
\end{split}
\end{equation*}
where ${\mathbf{p}_i^{(v_1)}}$ and ${\mathbf{p}_i^{(v_2)}}$ represent the $i$th columns of ${\mathbf{P}^{(v_1)}}$ and ${\mathbf{P}^{(v_2)}}$, respectively.
\end{definition}

For any cluster assignment $i$  $\left( {1 \le i \le 3} \right)$ in ${\mathbf{P}^{(v_1)}}$, one positive cluster assignment pair and four negative cluster assignment pairs are produced for the two views $v_1$ and $v_2$. To illustrate the realizability of similarity alignment, we introduce the definition of strict alignment.

\begin{theorem} \label{th:alignment}
For $n_v$ given views of multiview data, ${L_c}$ in Eq. \eqref{eq:lossc} is minimized if $f$ is strictly aligned $\forall {v_1},{v_2} \in \{ 1,2,...{n_v}\} $ and ${v_1} \ne {v_2}$.
\end{theorem}


Theorem \ref{th:alignment} shows that a lower bound in Theorem \ref{th:bound} can be theoretically achieved when strict alignment is satisfied according to Definition \ref{def:alignment}. The proof of Theorem \ref{th:alignment} is given in the supplementary material. Strict alignment is an ideal case, which implies that each cluster assignment $i$ $\left( {1 \le i \le 3} \right)$ in ${\mathbf{P}^{(v)}}$ $\left( {1 \le v \le n_v} \right)$ satisfies the conditions in Definition \ref{def:alignment}. Specifically, the first condition affects a single positive cluster assignment pair while the other two conditions are imposed on the four negative cluster assignment pairs.

In the general case, designing an encoder that is strictly aligned for multiple views is an intractable problem. Let $\mathbf{\widetilde p}_i^{(v)}$ be the $i$th row of ${\mathbf{P}^{(v)}}$, i.e., the feature of the $i$th instance in the $v$th view. Considering the cosine similarity measure, the distance betweem two features $\mathbf{\widetilde p_i}^{(v_1)}$ and $\mathbf{\widetilde p}_i^{(v_2)}$ is measured as
\begin{equation}
d\left( {\mathbf{\widetilde p}_i^{({v_1})},\mathbf{\widetilde p}_i^{({v_2})}} \right) = \frac{{\left\langle {\mathbf{\widetilde p}_i^{({v_1})},\mathbf{\widetilde p}_i^{({v_2})}} \right\rangle }}{{\left\| {\mathbf{\widetilde p}_i^{({v_1})}} \right\|\left\| {\mathbf{\widetilde p}_i^{({v_2})}} \right\|}}
\end{equation}
where $\left\langle { \cdot , \cdot } \right\rangle $ is the dot product operator. The cosine similarity may be inaccurate when two instances of a sample in views $v_1$ and $v_2$ belong to different domains of multiview data , e.g., text and image pairs. In addition, the alignment sensitivity is insufficient when considering the similarity between two features $\mathbf{\widetilde p_i}^{(v_1)}$ and $\mathbf{\widetilde p}_i^{(v_2)}$ in Eq. \eqref{eq:similarity}. For the proposed CVCL method, ${\mathbf{P}^{(v)}}$ is theoretically invariant to all types of views. From the point of view of alignment, the alignment of the cluster assignments exhibits a stronger ability to perform invariant representation learning than that of the instance features in MVC.

\subsubsection{Complexity Analysis}
Let $m$ and $s$ denote the minibatch size and the maximum number of neurons in the hidden layers of the proposed network architecture, respectively. The complexity of the feedforward computation is $\mathcal{O}\left( {n_v}m{d_v}{s^{(r+1)}} + {n_v}m{d_v}K \right)$ in the fine-tuning phase. The complexities of the reconstruction loss, cross-view contrastive loss and cross-view consistency loss are $\mathcal{O}\left( {n_v}{d_v}m \right)$, $\mathcal{O}\left( {m^2}K{n_v}\left( {\left( {{n_v} - 1} \right) + {n_v}\left( {K - 1} \right)} \right) \right)$ and $\mathcal{O}\left( {n_v}K \right)$, respectively. Therefore, the overall complexity of the proposed CVCL method is ${t}\left( {{n_v}m{d_v}{s^{(r+1)}} + n_v^2{m^2}{K^2} + {n_v}m{d_v}K} \right)$, where $t$ is the maximum number of iterations in the pretraining and fine-tuning phases.

\section{Experiments}
\label{sec:exp}
In this section, we conduct extensive experiments to evaluate the performance of the proposed CVCL method. The source code for CVCL is implemented in Python 3.9. The source code is available at https://github.com/chenjie20/CVCL. All experiments are conducted on a Linux workstation with a GeForce RTX 2080 Ti GPU (11 GB caches), an Intel (R) Xeon (R) E5-2667 CPU and 256.0 GB of RAM.

\subsection{Experimental Settings}
\subsubsection{Datasets}
The proposed CVCL method is experimentally evaluated on seven publicly available multiview datasets. The MSRC-v1 dataset contains 210 scene recognition images belonging to 7 categories \cite{Winn2005MSRC}. Each image is described by 5 different types of features. The COIL-20 dataset is composed of 1,440 images belonging to 20 categories \cite{NeneCOIL20}. Each image is described by 3 different types of features. The Handwritten dataset consists of 2,000 handwritten images of digits from 0 to 9 \cite{Asuncion2007UCI}. Each image is described by 6 different types of features. The BDGP dataset contains 2,500 samples of Drosophila embryos \cite{Cai2012SR}. Each sample is represented by visual and textual features. The Scene-15 dataset consists of 4,485 scene images belonging to 15 classes \cite{Li2005BHM}. Each image is represented by 3 different types of features. The MNIST-USPS dataset contains 5,000 samples with two different styles of digital images \cite{Asuncion2007UCI}. The Fashion dataset contains 10, 000 images of products \cite{Xiao2017FM}. Each image is represented by three different styles.

\begin{table*}[!htbp]
\scriptsize
\setlength{\tabcolsep}{3pt}
\centering
\caption{Results of clustering performance comparisons conducted on all datasets.}
\label{tb:clustering:results}
\begin{tabular}{c|ccc|ccc|ccc|ccc|ccc|ccc|ccc}
\hline
\multirow{2}*{Methods} & \multicolumn{3}{c|}{MSRC-v1} &  \multicolumn{3}{c|}{COIL-20} & \multicolumn{3}{c|}{Handwritten} & \multicolumn{3}{c|}{BDGP} & \multicolumn{3}{c|}{Scene-15} & \multicolumn{3}{c|}{MNIST-USPS} & \multicolumn{3}{c}{Fashion} \\
\cline{2-22}
~ & ACC & NMI & Purity & ACC & NMI & purity & ACC & NMI & purity & ACC & NMI & purity & ACC & NMI & purity & ACC & NMI & purity & ACC & NMI & purity \\
\hline
BSVC & 78.57 & 68.04 & 78.57 & 80.21 & 84.75 & 80.47 & 75.35 & 74.07 & 75.35 & 53.68 & 32.42 & 54.32 & 38.05 & 38.85 & 42.08 & 67.98 & 74.43 & 72.34 & 60.32 & 64.91 & 63.84 \\
SC$_{\textbf{Agg}}$ & 82.71 & 72.52 & 82.71 & 73.13 & 78.46 & 73.89 & 79.85 & 82.62 & 83.35 & 68 & 55.71 & 70.72 & 38.13 & 39.31 & 44.76 & 89 & 77.12 & 89.18 & 98 & 94.8 & 97.56 \\
EE-IMVC & 85.81 & 73.76 & 85.81 & 75.73 & 83.52 & 75.76 & 89.3 & 81.07 & 89.3 & 88 & 71.76 & 87.76 & 39 & 33.02 & 40.27 & 76 & 68.04 & 76.48 & 84 & 79.53 & 84.45 \\
ASR & 91.9 & 84.75 & 91.9 & \underline{80.9} & \underline{87.6} & \underline{81.5} & 93.95 & 88.26 & 93.95 & 97.68 & 92.63 & 97.68 & 42.7 & 40.7 & 45.6 & 97.9 & 94.72 & 97.9 & 96.52 & 93.04 & 96.52 \\
DSIMVC & 79.05 & 69 & 79.05 & 65.55 & 72.51 & 66.67 & 87.2 & 80.39 & 87.2 & \underline{99.04} & \underline{96.86} & \underline{99.04} & 28.27 & 29.04 & 29.79 & 99.34 & 98.13 & 99.34 & 88.21 & 83.99 & 88.21 \\
DCP & 78.57 & 74.84 & 79.43 & 67.36 & 78.79 & 69.86 & 85.75 & 85.05 & 85.75 & 97.04 & 92.43 & 97.04 & 42.32 & 40.38 & 43.85 & 99.02 & 97.29 & 99.02 & 89.37 & 88.61 & 89.37 \\
DSMVC & 85.24 & 76.96 & 85.24 & 76.46 & 84.15 & 78.19 & \underline{96.8} & \underline{92.57} & \underline{96.8} & 75.8 & 61.39 & 75.8 & \underline{43 .48} & \underline{41.11} & \underline{45.92} & 96.34 & 94.27 & 96.34 & 89.63 & 86.81 & 89.63 \\
MFL & \underline{93.33} & \underline{86.51} & \underline{93.33} & 73.19 & 81.43 & 75.07 & 86.55 & 85.98 & 86.55 & 98.72 & 96.13 & 98.72 & 42.52 & 40.34 & 44.53 & \underline{99.66} & \underline{99.01} & \underline{99.66} & \underline{99.2} & \underline{98} & \underline{99.2} \\
CVCL & \textbf{97.62} & \textbf{94.98} & \textbf{97.62} & \textbf{84.65} & \textbf{88.89}& \textbf{85.07} & \textbf{97.35} & \textbf{94.05} & \textbf{97.35} & \textbf{99.2} & \textbf{97.29} & \textbf{99.2} & \textbf{44.59} & \textbf{42.17} & \textbf{47.36} & \textbf{99.7}& \textbf{99.13} & \textbf{99.7} & \textbf{99.31} & \textbf{98.21} & \textbf{99.31} \\
\hline
\end{tabular}
\end{table*}

\begin{table*}[!htbp]
\scriptsize
\setlength{\tabcolsep}{1pt}
\centering
\caption{Ablation study concerning the main components of the proposed CVCL method on all the datasets.}
\label{tb:ablation}
\begin{tabular}{c|ccc|ccc|ccc|ccc|ccc|ccc|ccc|ccc}
\hline
\multirow{2}*{Methods} & \multirow{2}*{${L_{pre}}$} & \multirow{2}*{${L_c}$} & \multirow{2}*{${L_a}$} & \multicolumn{3}{c|}{MSRC-v1} &  \multicolumn{3}{c|}{COIL-20} &  \multicolumn{3}{c|}{Handwritten}  & \multicolumn{3}{c|}{BDGP} & \multicolumn{3}{c|}{Scene-15} & \multicolumn{3}{c|}{MNIST-USPS} & \multicolumn{3}{c}{Fashion}\\
\cline{5-25}
~ & ~ & ~ & ~ & ACC & NMI & purity & ACC & NMI & purity & ACC & NMI & purity & ACC & NMI & purity & ACC & NMI & purity & ACC & NMI & purity & ACC & NMI & purity \\
\hline
CVCL$_{\textbf{fine-tuning}}$ & ~ & \checkmark & \checkmark & 82.38 & 76.17 & 82.38 & 66.94 & 78.96 & 69.86 & 88 & 87.87 & 88 & 99.12 & 96.88 & 99.12 & 40.42 & 39.8 & 43.81 & 99.48 & 98.5 & 99.48 & 99.16 & 97.87 & 99.16 \\
CVCL$_{\textbf{Lc}}$ & \checkmark & \checkmark & ~ & 55.34 & 50.71 & 55.34 & 51.11 & 70.61 & 51.39 & 87.35 & 90.29 & 87.7 & 98.92 & 96.3 & 98.92 & 25.04 & 30.91 & 25.04 & 59.84 & 81.57 & 59.84 & 99.2 & 97.98 & 99.2 \\
CVCL & \checkmark & \checkmark & \checkmark & \textbf{97.62} & \textbf{94.98} & \textbf{97.62} & \textbf{84.65} & \textbf{88.89}& \textbf{85.07} & \textbf{97.35} & \textbf{94.05} & \textbf{97.35} & \textbf{99.2} & \textbf{97.29} & \textbf{99.2} & \textbf{44.59} & \textbf{42.17} & \textbf{47.36} & \textbf{99.7}& \textbf{99.13} & \textbf{99.7} & \textbf{99.31} & \textbf{98.21} & \textbf{99.31} \\
\hline
\end{tabular}
\vspace{-0.3cm}
\end{table*}

\subsubsection{Comparison Methods}
To validate the superiority of the proposed CVCL method, we compare CVCL with several state-of-the-art methods, including the efficient and effective incomplete MVC (EE-IMVC) algorithm, augmented sparse representation (ASR) algorithm \cite{Chen2022ASR}, deep safe IMVC (DSIMVC) algorithm \cite{Tang2022DSIMVC}, dual contrastive prediction (DCP) algorithm \cite{Lin2022DCP}, deep safe MVC (DSMVC) algorithm \cite{Tang2022DSMVC} and MFL \cite{Xu2022MLFL}. For DCP, the best clustering result is reported from the combinations of each pair of individual views in each dataset. In addition, two extra baselines are included for comparison purposes. Specifically, we first apply spectral clustering \cite{Luxburg2007SC} on each individual view and report the best clustering result obtained among multiple views, i.e., the best single-view clustering (BSVC) method. Then, we apply an adaptive neighbor graph learning method \cite{Nie2014CPC} to produce a similarity matrix for each individual view. We aggregate all similarity matrices into an accumulated similarity matrix for spectral clustering, which is referred to as SC$_{\textbf{Agg}}$.

\subsubsection{Evaluation Metrics}
Three widely used metrics are employed to evaluate the clustering performance of all competing algorithms, including the clustering accuracy (ACC), normalized mutual information (NMI), and purity \cite{Chen2021LRR}. For example, ACC considers the best matching result between two assignments, i.e., a cluster assignment obtained from an MVC algorithm and a known ground-truth assignment. For these metrics, a larger value indicates better clustering performance.

\subsubsection{Network Architecture and Parameter Settings}
\label{sec:settings}
The proposed network architecture consists of an input layer, hidden layers in the view-specific encoders, an extra linear layer and a softmax layer. A set of view-specific autoencoders is contained in the pretraining stage. The number of hidden layers possessed by the view-specific encoders or decoders ranges from 3 to 5. For example, the sizes of the 5 hidden layers are set to $\left[ {{d_v}, 256, 512, 1024, 2048, r_1} \right]$, where ${d_v}$ is the dimensionality of an instance in the $v$th view and $r_1$ is the dimensionality of the corresponding feature. Moreover, $r_2$ represents the size of the extra linear layer. We choose $r_1$ and $r_2$ from $\left[ {2000, 1500, 1024, 1000, 768, 512, 500, 256, 200} \right]$. The overall loss of the proposed network architecture has two parameters, $\alpha $ and $\beta $, which are chosen from $\left\{ 0.005, 0.01, 0.05 \right\}$ with a grid search strategy. For a fair comparison, the best clustering results of these competing methods are obtained by tuning their parameters.

\subsection{Performance Evaluation}
\label{sec:evaluation}
The clustering results produced by all competing methods on the seven multiview datasets are reported in Table \ref{tb:clustering:results}. The best and second-best values of the clustering results are highlighted in bold and underlined, respectively. The contrastive learning-based methods, including CVCL, DSMVC, DSIMVC and MFL, often achieve significant improvements over the other methods on large-scale datasets, e.g., the BDGP, MNIST-USPS, Scene-15 and Fashion datasets. Moreover, the CVCL method significantly outperforms the other contrastive learning-based methods, including DSMVC, DCP, DSIMVC and MFL, on all datasets. This verifies the importance of the cluster-level CVCL strategy. These results demonstrate the effectiveness of our proposed CVCL method. The proposed CVCL method achieves the best clustering results on all datasets. This is consistent with our theoretical analysis. For example, the CVCL method achieves performance improvements of approximately 4.29\%, 8.47\%, and 4.29\% over the second-best method on the MSRC-v1 dataset in terms of the ACC, NMI, and purity metrics, respectively. Similarly, the CVCL method performs much better than the other competing methods on the other datasets. These results demonstrate the superiority of CVCL over the other methods.

Two reasons explain the advantages and effectiveness of the proposed CVCL method. First, contrastive learning-based methods, e.g., CVCL, DSMVC, DSIMVC and MFL, consider deep representations for multiple views. We observe that they often achieve significant improvements over the other traditional methods, especially on large-scale datasets. Second, the alignment of soft cluster assignments plays a critical role in contrastive learning. By contrasting the cluster assignments among multiple views, the proposed CVCL method is guided to learn view-invariant representations in unsupervised learning. Consequently, it learns more discriminative view-invariant representations than the other contrastive learning-based methods.

\subsection{Ablation Studies}
According to the overall reconstruction loss in Eq. \eqref{eq:lossfine}, three different loss components are included. To verify the importance of each component in CVCL, we perform ablation studies with the same experimental settings to isolate the necessity of each component. Specifically, we consider two special cases: performing MVC in the fine-tuning stage without pretraining and performing MVC in both stages without the regularization term of the overall reconstruction loss ${L_a}$. These versions are referred to as CVCL$_{\textbf{fine-tuning}}$ and CVCL$_{\textbf{Lc}}$, respectively.

Table \ref{tb:ablation} shows the obtained clustering results in terms of the three metrics produced with the combinations of different loss components. The clustering results in the first two rows of Table \ref{tb:ablation} are achieved by the two special cases. As expected, the best performance can be achieved when all loss terms are considered and when the two-stage training scheme is employed in CVCL. Moreover, we can observe that the clustering performance is significantly improved when the pretraining stage is employed in CVCL. For example, CVCL performs much better than CVCL$_{\textbf{fine-tuning}}$, with improvements of approximately 5.24\%, 18.81\% and 15.24\% in terms of the ACC, NMI and purity metrics, respectively, achieved on the MSRC-v1 dataset. However, the clustering performance gap narrows as the number of samples significantly increases. For example, CVCL achieves 0.08\%, 0.22\% and 0.15\% ACC improvements over CVCL$_{\textbf{fine-tuning}}$ on the BDGP, MNIST-USPS and Fashion datasets, respectively. This indicates that an increase in the number of samples may reduce the significant advantages provided by the pretraining stage. In addition, the clustering performance achieved on most datasets dramatically declines when ${L_a}$ is ignored in the overall reconstruction loss. This indicates that   effectively guarantees that all instances can be assigned into clusters. Therefore, each component in the overall reconstruction loss plays a crucial role in learning view-invariant representations.

\begin{figure}
    \centering
    \subfigure[MSRC-v1]{
    \label{fig:MSRCv1:acc} 
    \includegraphics[width=3cm]{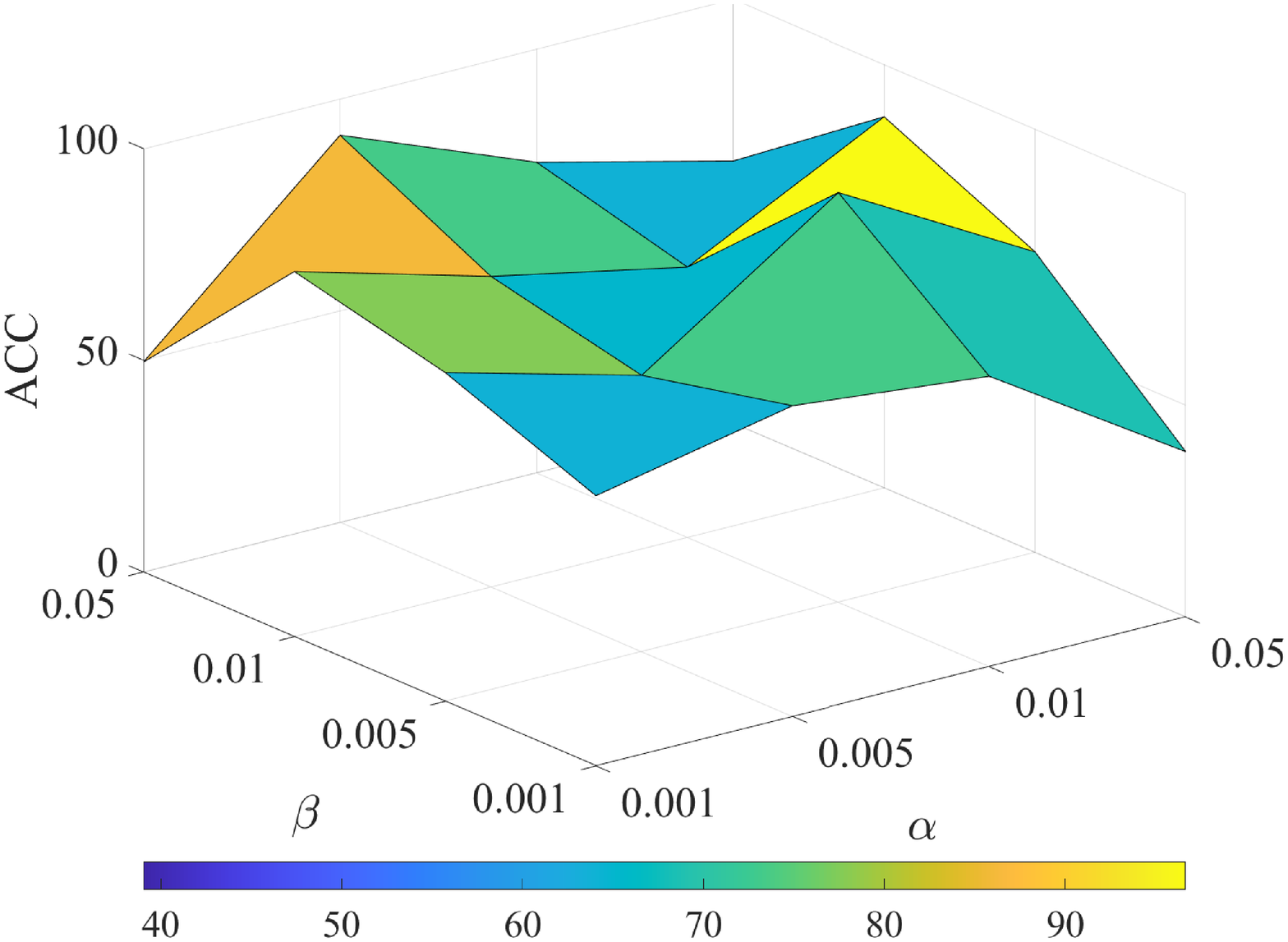}}
    \subfigure[Fashion]{
    \label{fig:fashion:acc} 
    \includegraphics[width=3cm]{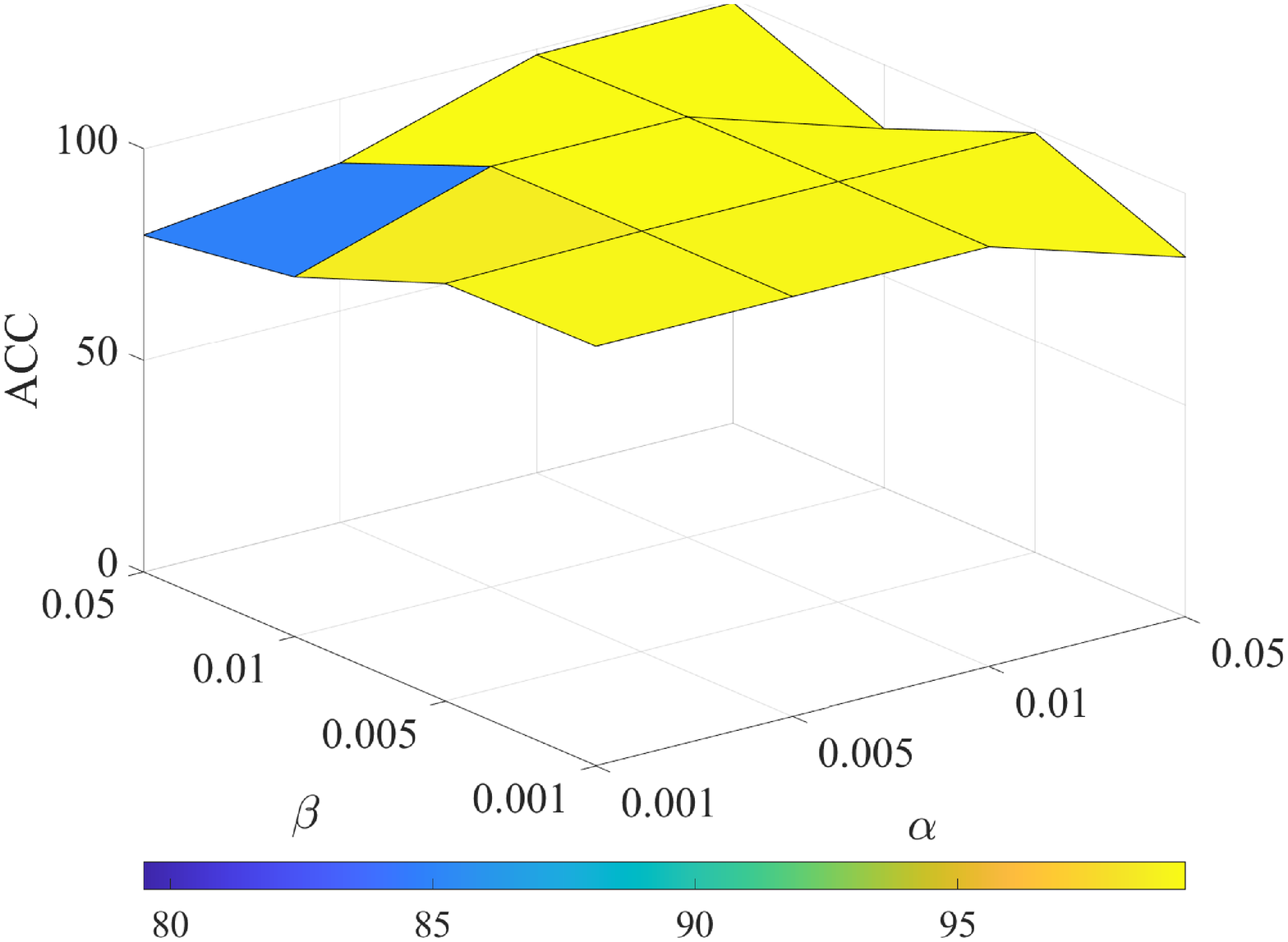}}
    \caption{The ACC values yielded by the CVCL method with different $\alpha$ and $\beta$ combinations on the four representative datasets.}
    \label{fig:sensitivity:acc} 
    \vspace{-0.5cm}
\end{figure}

\begin{figure}
    \centering
    \subfigure[MSRC-v1]{
    \label{fig:MSRCv1:nmi} 
    \includegraphics[width=3cm]{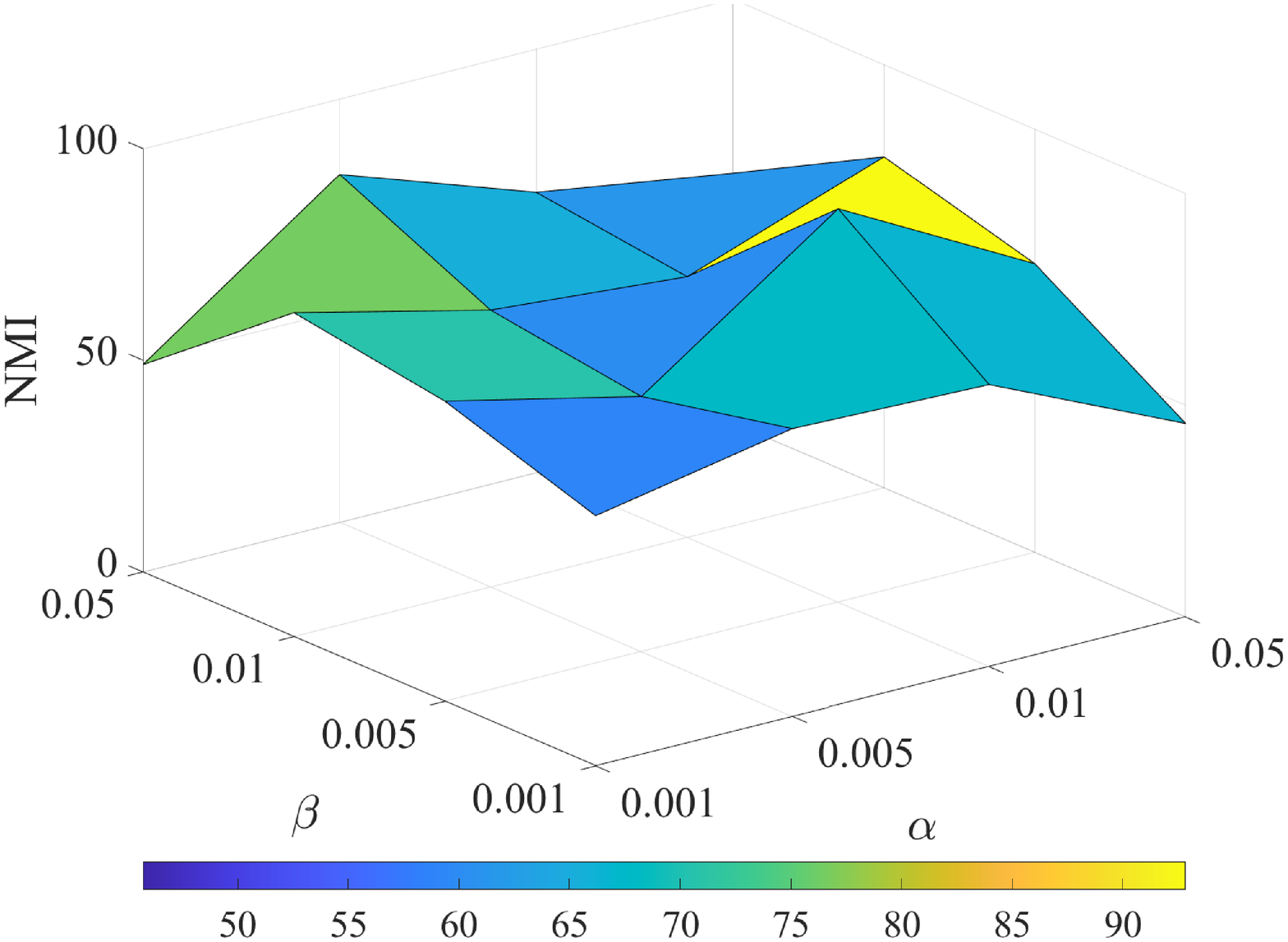}}
    \subfigure[Fashion]{
    \label{fig:fashion:nmi} 
    \includegraphics[width=3cm]{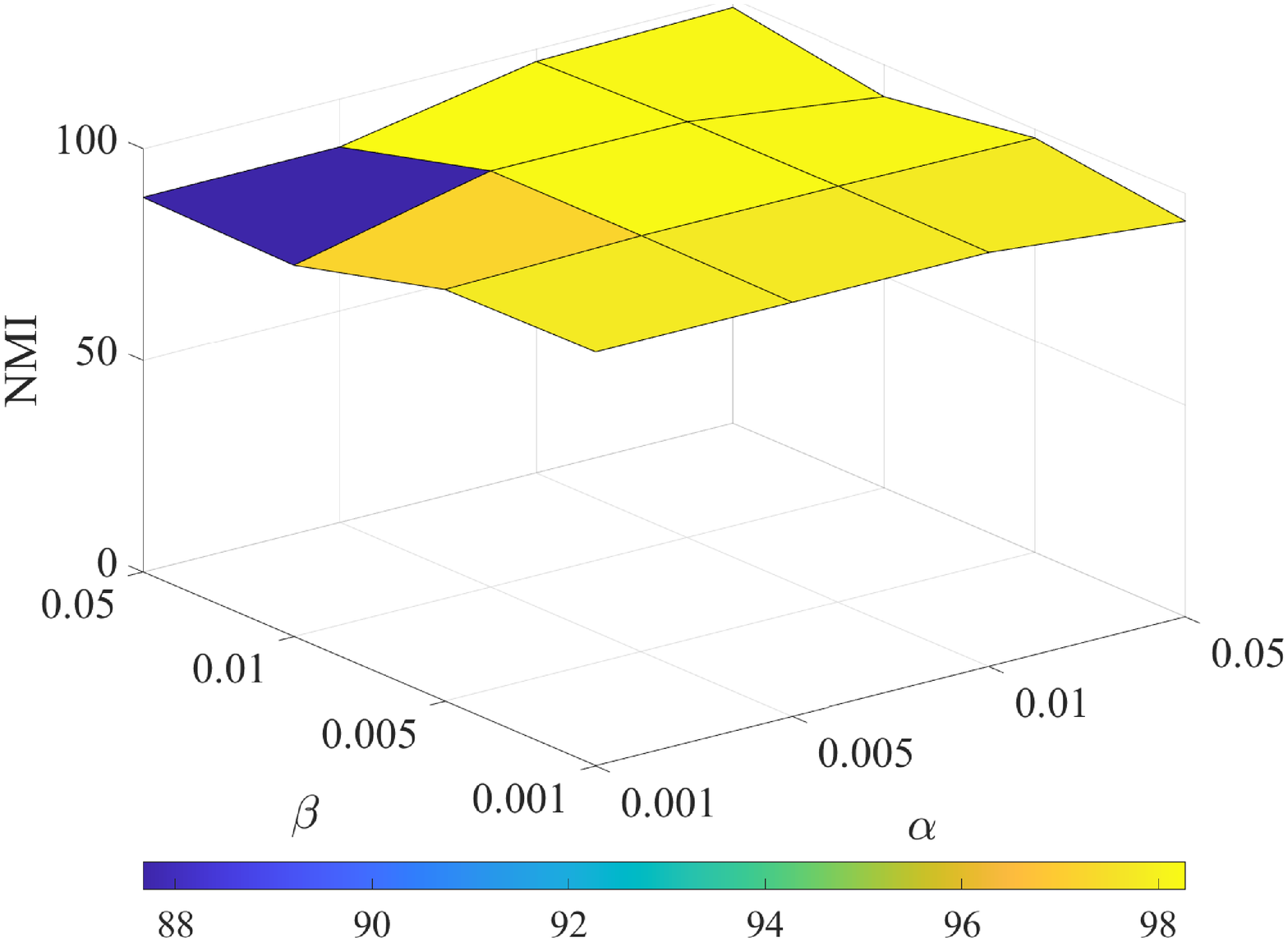}}
    \caption{The NMI values yielded by the CVCL method with different $\alpha$ and $\beta$ combinations on the two representative datasets.}
    \label{fig:sensitivity:nmi} 
    \vspace{-0.5cm}
\end{figure}

\begin{figure}
\centering
\subfigure[The pretraining stage]{
\label{fig:conv:a} 
\includegraphics[width=3cm]{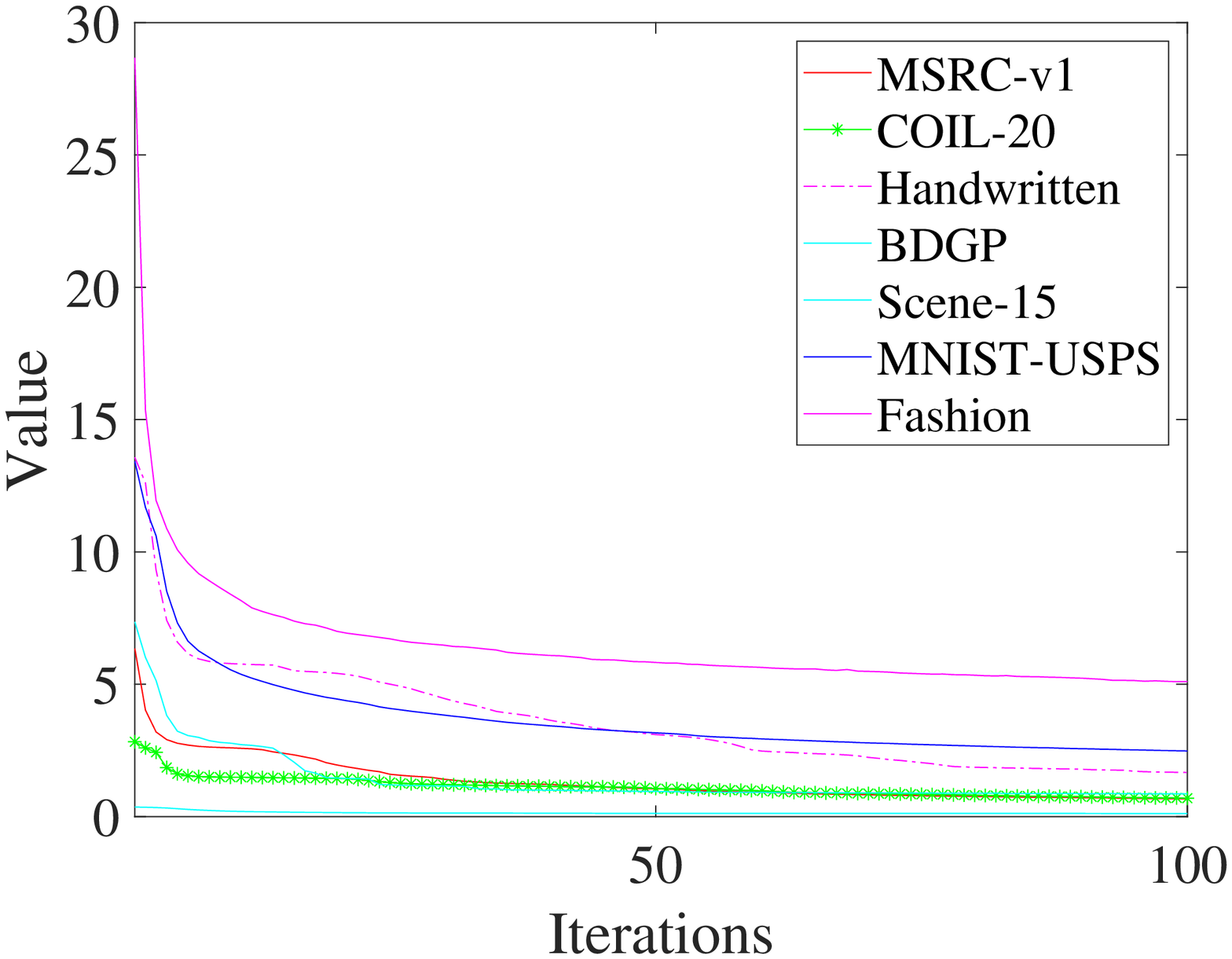}}
\hfil
\subfigure[The fine-tuning stage]{
\label{fig:conv:b} 
\includegraphics[width=3cm]{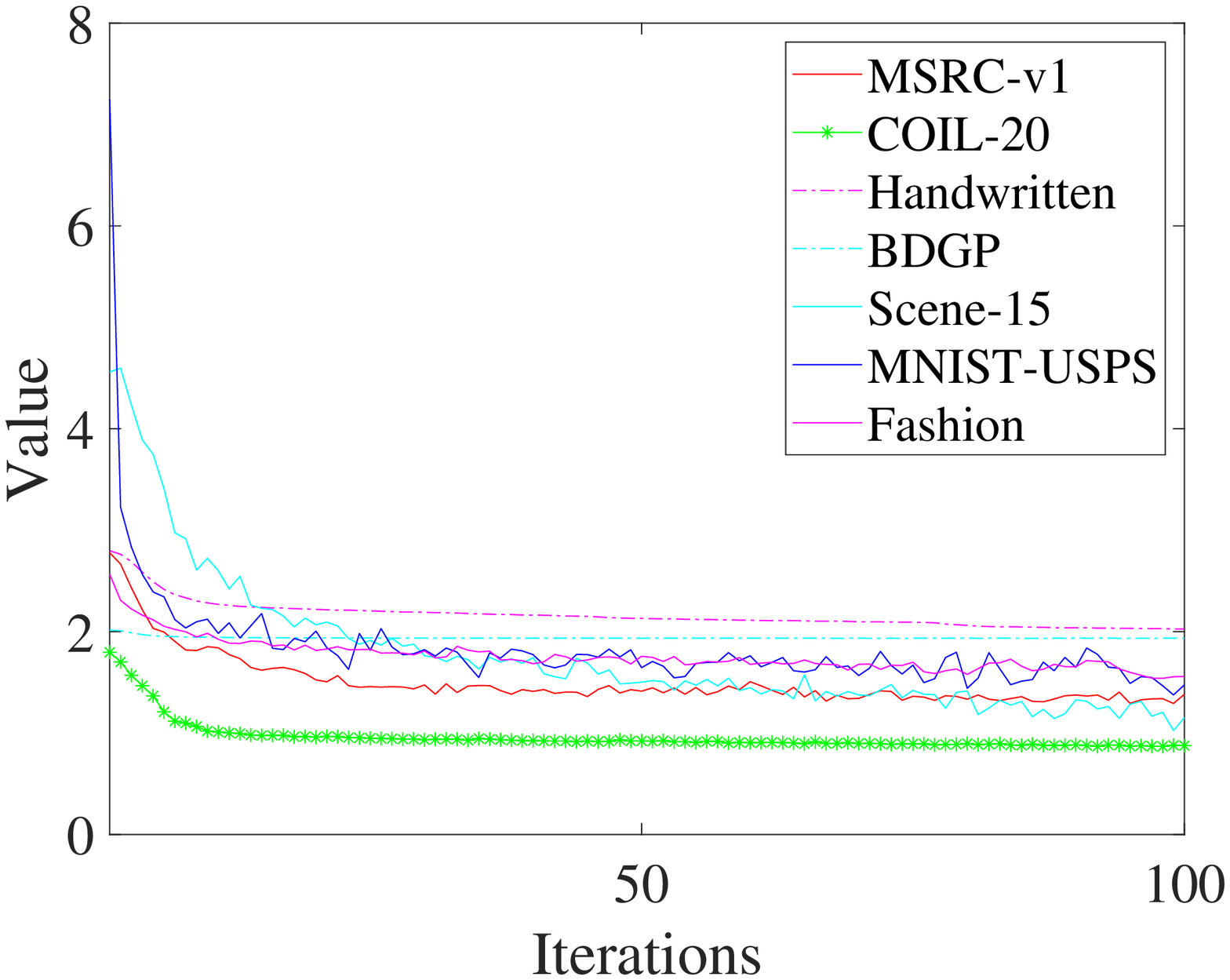}}
\hfil
\caption{Convergence results obtained by the CVCL method on all the datasets.}
\label{fig:conv} 
\vspace{-0.5cm}
\end{figure}

\subsection{Parameter Sensitivity Analysis}
We conduct experiments on two representative datasets, i.e., the MSRC-v1 and Fashion datasets, to investigate the sensitivity of the $\alpha $ and $\beta $ parameters in the proposed CVCL method. The $\alpha $ and $\beta $ parameters are chosen from $\left\{ 0.001, 0.005, 0.01, 0.05 \right\}$ for CVCL. Figures \ref{fig:sensitivity:acc} and \ref{fig:sensitivity:nmi} show the clustering performance achieved by the CVCL method in terms of the ACC and NMI values obtained with different combinations of $\alpha $ and $\beta$. It can be observed that the clustering performance attained by the CVCL method on the MSRC-v1 dataset seriously fluctuates with different combinations of $\alpha $ and $\beta$. As the number of samples dramatically increases in the other dataset, the CVCL method can achieve relatively stable clustering results with most combinations of $\alpha $ and $\beta$. This indicates that the CVCL method has stable clustering performance when utilizing a larger number of samples.

\subsection{Training Analysis}
We investigate the convergence of the CVCL method. Two major learning stages are contained in the CVCL method, including the pretraining and fine-tuning stages. To validate the convergence of the CVCL method, we compute the results of the loss functions in Eqs. \eqref{eq:lossr} and \eqref{eq:lossfine} during these two stages. Figure \ref{fig:conv} shows the curves of the loss function results obtained on all the datasets. The values of the loss function in Eq. \eqref{eq:lossr} dramatically drop in the first few iterations and then slowly decrease until convergence is achieved. We also observe a similar trend in the changes in the loss function values in Eq. \eqref{eq:lossfine} on most datasets, e.g., the COIL-20, Handwritten and BDGP datasets. In addition, the curves of the loss function in Eq. \eqref{eq:lossfine} produced on the other datasets fluctuate slightly after the first few iterations. These results demonstrate the effectiveness of the convergence property of the CVCL method.

\section{Conclusion}
\label{sec:conclusion}
In this paper, we propose a CVCL method that learns view-invariant representations for MVC. A cluster-level CVCL strategy is presented to explore the consistent semantic label information possessed among multiple views.CVCL effectively achieves more discriminative cluster assignments during two successive stages. A theoretical analysis of soft cluster assignment alignment indicates the importance of the cluster-level learning strategy in CVCL. We conduct extensive experiments and ablation studies on MVC datasets to validate the superiority of the model and the effectiveness of each component in the overall reconstruction loss.

\section*{Acknowledgements}
This work was supported by National Natural Science Foundation of China (NSFC) under Grant 62176171 and Grant U21B2040.

{\small
\bibliographystyle{ieee_fullname}
\bibliography{cvcl}
}

\end{document}